# GRAVITATIONAL CELL DETECTION AND TRACKING IN FLUORESCENCE MICROSCOPY DATA

*Nikomidisz Jorgosz Eftimiu, Michal Kozubek*

Centre for Biomedical Image Analysis, Faculty of Informatics, Masaryk University, Brno, Czech Rep.
eftimiu@mail.muni.cz, kozubek@fi.muni.cz

**ABSTRACT**

Automatic detection and tracking of cells in microscopy images are major applications of computer vision technologies in both biomedical research and clinical practice. Though machine learning methods are increasingly common in these fields, classical algorithms still offer significant advantages for both tasks, including better explainability, faster computation, lower hardware requirements and more consistent performance. In this paper, we present a novel approach based on gravitational force fields that can compete with, and potentially outperform modern machine learning models when applied to fluorescence microscopy images. This method includes detection, segmentation, and tracking elements, with the results demonstrated on a Cell Tracking Challenge dataset.

## 1. INTRODUCTION

In modern clinical practice and especially biomedical research, automatic cell tracking and analysis can significantly decrease the workload on human experts and guarantee consistent quality of results. With the continued evolution of computer hardware and software, an increasing number of academic and commercial projects have focused on creating end-to-end pipelines for solving these tasks without human intervention. Though statistical and machine learning approaches are especially popular nowadays due to their ability to model complex phenomena without explicit programming, classical algorithms remain widespread in the field of image analysis, largely due to their better explainability, extensibility and lower consumption of resources (time, memory, CPU/GPU, energy, etc.). Not only can these methods be modified and applied to novel problems with ease, but they can often still compete with modern deep learning systems, as the results of the Cell Tracking Challenge (CTC) [1] show.

In this paper, we present an algorithm for solving the above-mentioned detection and tracking tasks using a lightweight, classical algorithm utilising a physics-based, gravitational model. In the following sections, we describe the principles behind the algorithm's operation, then showcase the results obtained on one of the CTC fluorescence datasets using this pipeline.

## 2. PRIOR WORK

The concept of using physics-based models for image segmentation is not wholly unknown, and several examples of successful projects exist in literature. Perhaps the most notable among them is the work of Hurley et al. [2], who employed gravitational force field transforms for shape description in ear biometrics. Nixon et al. [3], continuing the work of Hurley, proposed several physical models for segmenting MRI and CT images, specifically water flow- and heat transfer-based algorithms. Gravitational approaches also appear in the works of Rashedi et al. [4] and Chao et al. [5], who applied their methods to light microscopy images of semiconductors and generic RGB images, respectively. Similarly to Chao et al., Upadhyay et al. [6] also used gravitational forces to perform colour clustering in RGB images with the help of search agents. Deregeh et al. [7] also used an agent-based approach to perform edge detection in grayscale images, with their algorithm seemingly outperforming many classical techniques.

Despite the concept of using gravity for image segmentation not being entirely novel, very few projects have attempted to use this method for locating objects of interest, especially in biomedical data. As such, gravitational methods have not been adopted by the wider computer vision community, and these approaches remain relatively unexplored.

## 3. METHODOLOGY

**3.1 Preprocessing**

Prior to applying the gravity method, it is necessary to preprocess the images to enhance weak features and reduce noise. As such, the individual frames are first brightened using a log transform, then they are filtered. In order to preserve thin, linear structures and reduce the possibility of cells merging together, an edge-aware filter, the Kuwahara filter [8] was chosen for this task. This algorithm was originally developed for biomedical applications, and in our project, we employed a high performance, anisotropic version developed by Kyprianidis et al. [9]. A key advantage of this approach over other common edge-aware filtering strategies (such as Perona-Malik diffusion) is its non-iterative nature, which reduces its computational cost compared to alternatives.

### 3.2 Detection

After preprocessing, the images are used to generate gravitational force fields using Newton's law of gravitation. In this step, every pixel within the frame is treated as a point-like body with a mass equal to its brightness, and the forces arising at a given point are calculated by summing the contributions of individual pixels, as seen in Fig. (1). This operation can be implemented efficiently as a convolution with a suitable kernel, calculated using the inverse distances from a central location. For a completely valid implementation, it would be necessary to employ a kernel that covers the entire image at every location, though this was found to significantly degrade the accuracy of the detection, mostly due to the gravitational fields of brighter cells overpowering those of the darker ones. As such, kernels with sizes comparable to the expected sizes of the cells were chosen, with reflection padding around the borders of the image to ensure that even partial cells would be detected.

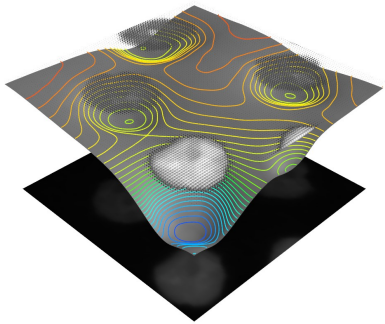

**Fig. 1.** An example potential field generated from the image shown at the bottom. Every pixel is treated as a point mass, creating the potential surface and contour lines shown above.

Cell candidates are detected by determining the positions of local minima within the potential fields of the images, though a direct approach is problematic for several reasons. Most notably, the potential fields have many spurious minima, especially far away from any cells where the gravitational field is relatively weak, and it is impossible to determine which of the minima are significant. To remedy this, the gravitational force (Formula (1)) is used to perform gradient descent and ascent via numerical integration.

To filter out spurious minima, the basins of attraction are extracted from the force fields. These are the regions of the image where forces arising from one object dominate over all others; that is to say, if a state is initialized anywhere within said basin during gradient descent, it will converge to the same minimum without exception. This definition is equivalent to a watershed transform by the drop of water principle, i.e., that catchment basins are those regions where a drop of water will flow towards the same minimum regardless of where it is initialized from.

$$a = \sum_{i=0}^{n} -\frac{GM}{(x_i - r)^2} \hat{x}_i$$

**Formula. 1.** The force field is calculated by summing the contributions of every pixel, scaled by the square of the distance. G and M denote the gravitational constant and mass, respectively. x is a vector pointing towards a pixel, x̂ is a unit vector in the same direction and r is the distance.

In practice, the watershed transform could not be used due to implementation issues, so a different, equivalent definition was used to extract the basins of attraction. First, a continuous force field is approximated from the discrete version using bilinear interpolation, then points where both components of the force field are zero, the critical points of the system, are located. The partial differential equations governing the gradient descent are then linearized using the Jacobian matrix, and saddle points are localized. After this, segmenting the image into basins of attraction is a simple matter of performing gradient ascent from the saddles to follow the stable manifolds of the force field, which delimit said basins. To ensure accurate results, a numerical integration scheme with adaptive step size based on the Euler-Heun method is used. The coefficients were reused from a technical report by E. Fehlberg [10] and can be seen in Table (1).

| | | | |
|---|---|---|---|
| 0 | | | |
| 1 | 1 | | |
| 1/2 | 1/4 | 1/4 | |
| | 1/6 | 1/6 | 4/6 |
| | 1/2 | 1/2 | |

**Table 1.** The extended Butcher tableau defining the adaptive numerical integration method.

Minima that correspond to basins of sufficiently large size are treated as significant cell candidates, while all other minima are discarded. This approach favours large, bright objects, especially ones surrounded by background regions that are darker than their interiors. Such high contrast objects often present one or only a few minima with basins of attraction comparable in size to the objects themselves, while low intensity background regions tend to contain many minima in close proximity to one another.

One major advantage of this method is its insensitivity to noise, since the convolution operation is inherently robust against small variations in image intensity. This averaging effect ensures that even cells with non-uniform brightness are located consistently, although filtering the image is still necessary for the subsequent segmentation step.

### 3.3 Segmentation

Once candidate locations are extracted from the image, the cell bodies must be segmented. This step is performed by first enhancing the original image using contrast limited adaptive histogram equalization (CLAHE), then applying morphological reconstruction to fill in any dark spots within the cytoplasm, which can be caused, for example, by unstained nuclei. Afterwards, a region growing algorithm is applied, which propagates the cells masks outward from the gravitational minima using local contrast as a delimiting factor. This approach is similar to how "magic wand" selection tools found in many image manipulation programs operate. In practice, the pixels that contain minima are inserted into a heap data structure based on their value, and they are examined in the order of brightest-to-darkest. If a pixel is darker than the average of the pixels currently contained in the mask minus a given threshold, it is marked as a wall; otherwise, the pixel is added to the mask, all of its non-wall neighbours are inserted into the heap and the process repeats until termination. The threshold for marking the pixel as a wall is based on the average value of the mask, and is therefore different for bright and dark cells.

After extraction, the cell masks are refined using a modified, distributional Chan-Vese algorithm based on the work of Cremers et al. [11]. This method can accurately locate the boundaries of cytoplasm regions, though it is prone to segmenting touching cells as a single instance, which necessitated the development of a splitting strategy. Since the cells that we analysed can be assumed to be mostly circular in shape with no significant protrusions or elongation, a watershed transform was employed for this purpose. The seed points of the algorithm were defined as the h-maxima of a distance transform applied to each mask separately.

### 3.4 Tracking

To perform tracking, the basins extracted in the detection step are re-used to associate the cells along the temporal axis. Since a single mask may contain multiple gravitational minima however, the complete attraction basins corresponding to given cell instances must be recovered. This is accomplished by re-labelling the original basins using the output of the segmentation step to merge the image regions that all attract to the same cell. An example of the complete pipeline's operation can be seen in Fig. (2).

In a three-stage process, the masks from the segmentation step are compared against the following and the previous frames, first in a backward, then a forward pass, followed by a final, linking pass. This method allows the algorithm to locate missing detections and construct the complete tracklets with high accuracy.

To associate the cells in neighbouring frames, every pixel within a given mask is attracted to minima within the next or previous image. Since the basins from the detection phase are labelled to represent cell instances, this operation can be performed in near-constant time. In the vast majority of cases, a cell will map to its own future or past gravitational field, and thus create a one-to-one matching with its corresponding label.

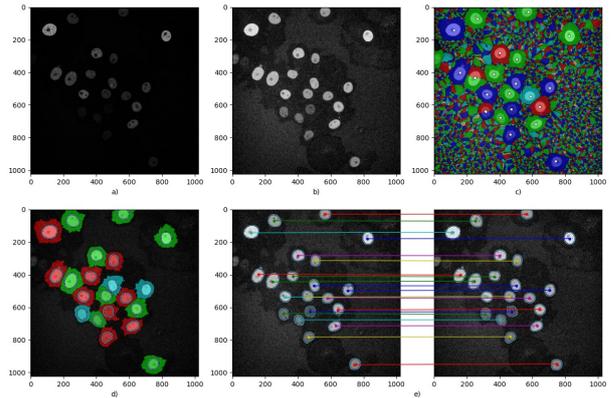

**Fig. 2.** An example of the full pipeline. **a)** Original image. **b)** Preprocessed image. **c)** The basins of attraction shown via equitable graph colouring and the significant minima corresponding to the detected cells denoted by white markers. **d)** The segmentations shown around the cells and the merged attraction basins of the cells. **e)** The tracking results for the current and the next images.

If no match is found for a given cell, then the algorithm can still attempt to find a missing detection by segmenting regions near its original location. In this case, Chan-Vese segmentation is performed using the previous cell mask, and the local contrast of the candidate cell is compared against the known value of the existing mask. If the proposed borders are too indistinct, then no new match is created, while clear foreground objects are added to the detections. Similarly, locations where cells are detected in both the previous and the next frame but not in the current one are unconditionally marked as detections. In this case, an interpolation method developed by Iwanowski [12] is used to create a smooth transition between the two end states.

After refining the initial detections, a final backward pass is used to construct the tracklets and re-label the masks to be consistent with CTC requirements. The reason for performing this step in reverse time is that, unlike in the forward direction, merging events can be allowed; if two cells join in a previous frame, that can be treated as a mitosis event and labelled appropriately.

This pipeline can often reduce the number of false negatives, though it also has a tendency to introduce additional false positives into the detections. Since no validation is performed to ensure that the masks added during detection are similar to their parents or descendants, errors can propagate far along the video sequence. To address this issue, the linked tracklets are examined once more to filter out obvious false detections. In the current version of the algorithm, a hysteresis thresholding system not unlike a Canny edge detector is used for this purpose. Considering one tracklet at a time, the sizes of the cell masks are compared against a lower and an upper bound, then the entire cell line is kept or discarded based on the result. If any mask in the tracklet is smaller than the lower bound or all masks are smaller than the upper bound, then they are treated as false detections, otherwise they are retained. Furthermore, the local contrast of the detected cells is examined once again to filter out any background regions that may have been incorrectly labelled as foreground objects.

## 4. EXPERIMENTAL RESULTS

The method was evaluated on the Fluo-N2DH-GOWT1 dataset of the Cell Tracking Challenge in accordance with the usual submission guidelines. The numerical results and performance measures were evaluated on the test datasets, while the algorithm's behaviour was examined using the gold-truth annotations of the training datasets. The parameters were tuned for the training datasets and adjusted after visual inspection of the test datasets, with the same parameter set being reused for all video sequences. The results can be seen in Table (2).

| Metric | Dataset 01 | Dataset 02 | Combined score | Ranking |
|---|---|---|---|---|
| DET | 0.966 | 0.973 | 0.970 | 5/51 |
| SEG | 0.879 | 0.919 | 0.899 | 11/41 |
| TRA | 0.960 | 0.970 | 0.965 | 4/41 |
| $OP_{CSB}$ | 0.923 | 0.946 | 0.934 | 8/51 |
| $OP_{CTB}$ | 0.920 | 0.945 | 0.932 | 5/41 |

**Table 2.** The results of the method on the Fluo-N2DH-GOWT1 dataset.

Compared to existing submissions in the CTC, our method achieves fourth and fifth place in terms of TRA score and DET score, respectively. Though segmentation was not the primary aim of our project, which significantly reduced the SEG scores compared to other projects, the method still achieved fourth place on the tracking benchmark leaderboards. Furthermore, it is currently the second-best performing non-machine learning-based approach among CTC submissions for the Fluo-N2DH-GOWT1 dataset in terms of both DET and TRA measures.

Though the gold truth annotations are not made available for the test datasets, it is possible to draw some conclusions regarding the shortcomings of the model by examining its behaviour during operation. Concerning the errors seen on the test datasets, some typical failure cases have been identified. In most instances, the greatest source of penalties are false negatives, which often occur when two cells of different intensities are in close proximity to one another, and no gravitational minimum is present within one of the cells. Similarly, cell clumps usually share a single minimum and separation is only possible if all members of a cluster have regular, well-defined shapes.

Other especially common reasons for false negatives include mitosis events, since chromatin in cell nuclei tends to condense and cells tend to exit the focal plane during divisions. This causes the already thin, unusually-shaped structures of the cell to become almost indistinguishable from background artifacts, which in turn causes the failure of the detection or the segmentation step. This issue is further exacerbated by the presence of many out-of-focus cells, since they can prevent the region growing stage from terminating and create uneven cell masks that get filtered out during later stages.

In terms of the algorithm's advantages, one notable observation is that the method can locate even very faint and indistinct cells in background regions, many of which were not included in the original annotations of the training datasets.

In addition to the CTC evaluation, we also measured the running times of the algorithm to enable comparisons to other methods. On a desktop workstation equipped with an Intel Core i9-10900 CPU, NVIDIA GeForce RTX 2060 SUPER GPU and 16 GB of memory, the basin extraction and detection stages took around 1 second each per 1024x1024 image, though these stages could be parallelized to decrease processing times. The tracking and filtering steps, which are inherently sequential, took a further 2 seconds and 0.5 seconds, respectively. As such, processing an entire sequence of 92 images took 6-7 minutes on average.

## 5. FUTURE WORK

Though the method already gives excellent results, numerous refinements could be made to the pipeline, including accurate handling of mitosis events and more sophisticated filtering of the tracklets. These additions could all enhance the performance of the model and possibly improve the CTC metrics by a significant margin. Furthermore, some parts of the algorithm could be simplified and accelerated, most notably the basin extraction step, and our current focus is implementing a custom watershed algorithm to replace the numerical integration method. In addition to improvements to the model, we also plan to make a second submission to the CTC once the algorithm is general enough to be applied to other fluorescence datasets.

## 6. ACKNOWLEDGEMENTS

The authors declare no conflict of interest, financial or otherwise. This research study was conducted retrospectively using human subject data made available in open access. Ethical approval was *not* required as confirmed by the license attached with the open access data. The work was supported by the Ministry of Education, Youth and Sports of the Czech Republic (Project LM2023050).

## 7. REFERENCES

[1] M. Maška et al., 'The Cell Tracking Challenge: 10 years of objective benchmarking', Nat Methods, vol. 20, no. 7, pp. 1010–1020, 2023, doi: 10.1038/s41592-023-01879-y.

[2] D. J. Hurley, M. S. Nixon, and J. N. Carter, 'Force field energy functionals for image feature extraction', Image Vis Comput, vol. 20, no. 5, pp. 311–317, 2002, doi: https://doi.org/10.1016/S0262-8856(02)00003-3.

[3] M. S. Nixon, X. U. Liu, C. Direkoğlu, and D. J. Hurley, 'On Using Physical Analogies for Feature and Shape Extraction in Computer Vision', Comput J, vol. 54, no. 1, pp. 11–25, Jan. 2011, doi: 10.1093/comjnl/bxp070.

[4] E. Rashedi and H. Nezamabadi-pour, 'A stochastic gravitational approach to feature based color image segmentation', Eng Appl Artif Intell, vol. 26, no. 4, pp. 1322–1332, 2013, doi: https://doi.org/10.1016/j.engappai.2012.10.002.

[5] Y. Chao, M. Dai, K. Chen, P. Chen, and Z. Zhang, 'A novel gravitational search algorithm for multilevel image segmentation and its application on semiconductor packages vision inspection', Optik (Stuttg), vol. 127, no. 14, pp. 5770–5782, 2016, doi: https://doi.org/10.1016/j.ijleo.2016.03.059.

[6] P. Upadhyay and J. K. Chhabra, 'An un-supervised image segmentation technique based on multi-objective Gravitational search algorithm (MOGSA)', in 2016 1st India International Conference on Information Processing (IICIP), 2016, pp. 1–4. doi: 10.1109/IICIP.2016.7975355.

[7] F. Deregeh and H. Nezamabadi-pour, 'A new gravitational image edge detection method using edge explorer agents', Nat Comput, vol. 13, no. 1, pp. 65–78, 2014, doi: 10.1007/s11047-013-9382-9.

[8] M. Kuwahara, K. Hachimura, S. Eiho, and M. Kinoshita, 'Processing of RI-Angiocardiographic Images', in Digital Processing of Biomedical Images, K. Preston and M. Onoe, Eds., Boston, MA: Springer US, 1976, pp. 187–202. doi: 10.1007/978-1-4684-0769-3_13.

[9] J. Kyprianidis, A. Semmo, H. Kang, and J. Döllner, Anisotropic Kuwahara Filtering with Polynomial Weighting Functions. 2010. doi: 10.2312/LocalChapterEvents/TPCG/TPCG10/025-030.

[10] E. Fehlberg and G. C. Marshall, 'Low-order classical Runge-Kutta formulas with stepsize control and their application to some heat transfer problems', 1969. [Online]. Available: https://api.semanticscholar.org/CorpusID:117960134

[11] D. Cremers, O. Fluck, M. Rousson, and S. Aharon, 'A probabilistic level set formulation for interactive organ segmentation', in Proc.SPIE, Mar. 2007, p. 65120V. doi: 10.1117/12.708609.

[12] M. Iwanowski, 'Morphological normalized binary object metamorphosis', vol. 32, 2006, pp. 626–632. doi: 10.1007/1-4020-4179-9_90.